\documentclass[runningheads]{llncs}

\usepackage{eccv}

\usepackage{eccvabbrv}

\usepackage{graphicx}
\usepackage{booktabs}
\usepackage{wrapfig}
\usepackage{caption}
\usepackage[accsupp]{axessibility} 

\usepackage{booktabs}
\usepackage{multirow}
\usepackage[table]{xcolor}
\usepackage{subcaption}
\definecolor{lightb}{RGB}{225, 240, 255} 
\definecolor{darkb}{RGB}{190, 220, 255}
\definecolor{lightr}{RGB}{255, 225, 225}
\usepackage{orcidlink}
\nocite{*}

\begin{document}

\title{LA-Sign: Looped Transformers with Geometry-aware Alignment for Skeleton-based \\ Sign Language Recognition} 

\titlerunning{Abbreviated paper title}

% Second Author\inst{2,3}\orcidlink{1111-2222-3333-4444} \and
% Third Author\inst{3}\orcidlink{2222--3333-4444-5555}}

\author{
Muxin Pu\inst{1} \and
Mei Kuan Lim\inst{1} \and
Chun Yong Chong\inst{1} \and
Chen Change Loy\inst{2}
}

\institute{
School of Information Technology, Monash University \and
S-Lab, Nanyang Technological University \\
\email{muxin.pu@monash.edu}
}


% Springer Heidelberg, Tiergartenstr.~17, 69121 Heidelberg, Germany
% ABC Institute, Rupert-Karls-University Heidelberg, Heidelberg, Germany

\maketitle 

\begin{abstract}
Skeleton-based isolated sign language recognition (ISLR) demands fine-grained understanding of articulated motion across multiple spatial scales, from subtle finger movements to global body dynamics. Existing approaches typically rely on deep feed-forward architectures, which increase model capacity but lack mechanisms for recurrent refinement and structured representation. We propose \textbf{LA-Sign}, a looped transformer framework with geometry-aware alignment for ISLR. Instead of stacking deeper layers, LA-Sign derives its depth from recurrence, repeatedly revisiting latent representations to progressively refine motion understanding under shared parameters. To further regularise this refinement process, we extend static geometry-aware alignment into a progressive training signal at each looping iteration, providing explicit geometric guidance to intermediate latent representations throughout the refinement process rather than as a static objective applied once at the final step. This encourages multi-scale semantic organisation within an adaptive hyperbolic space. We study three looping designs and multiple geometric manifolds, demonstrating that encoder-decoder looping combined with adaptive Poincaré alignment yields the strongest performance. Extensive experiments on WLASL and MSASL benchmarks show that LA-Sign achieves state-of-the-art results while using fewer unique layers, highlighting the effectiveness of recurrent latent refinement and geometry-aware representation learning for sign language recognition. The artefacts and code related to this study are made publicly online. \footnote{https://github.com/mpuu00001/LA-Sign}
\end{abstract}

\section{Introduction}
\begin{quote}
“\textit{We do not learn from experience; we learn from reflecting on experience.}” --- John Dewey
\end{quote}

When a signer lifts their hand to begin a motion, meaning does not appear all at once. It unfolds gradually: a change in finger tension, a slight rotation of the wrist, or a shift of the arm that is almost imperceptible unless one returns to the motion repeatedly to discern what truly distinguishes one sign from another. This fine balance between subtle motion and semantic interpretation lies at the core of skeleton-based isolated sign language recognition (ISLR) \cite{zhang2016chinese, jiang2021skeleton, jiang2021sign, zuo2023natural, pu2024siformer, Li_2025_CVPR}, a fundamental task that aims to infer the meaning of sign language (SL) at the word level using compact and efficient skeletal representations.

In ISLR, understanding is shaped not only by what is seen but also by how it is considered. SLs evolve culturally, vary widely across communities, and lack the structural regularities \cite{emmorey2001language, senghas2004children, meir2010emerging, lutzenberger2023social} that often make brute-force scaling effective in other domains. When meaning is distributed across the hands, arms, and face in an ever-shifting hierarchy, the central challenge is not merely to process more data but to process it with greater depth and sensitivity.

This observation naturally invites a question: What if a model could pause, reconsider, and refine its own internal interpretation of a sign, much like a human observer replaying the motion to resolve ambiguity? Looped transformers offer a promising route toward this capability. Rather than increasing depth by stacking additional layers, looped transformers reuse the same framework and revisit their latent representations \cite{dehghani2018universal, lan2019albert, giannou2023looped, yang2023looped, csordas2024moeut, fan2024looped, saunshi2025reasoning, chen2025inner}, with each pass uncovering structure that the previous one may only have hinted at.

The strength of recurrent priors for learning complex problems has already been demonstrated in the “deep thinking” literature \cite{schwarzschild2021can, schwarzschild2021datasets, bansal2022end}. They show that such loops can emulate iterative algorithms and internal chains of reasoning at constant parameter cost, treating each loop as a “latent thought” that deepens interpretation without expanding the network. Empirical findings \cite{saunshi2025reasoning, gatmiry2024can} further reveal that looped models often match deeper non-looped transformers precisely because this iterative revisiting sharpens ambiguous internal states. Seen through the lens of SL, this training paradigm remains largely unexplored. In this context, looping is not merely a computational trick; it embodies the very cognitive act of reflecting on motion.

But reflection alone is not enough. As recent findings in hyperbolic learning suggest, Euclidean representations often distort the inherent hierarchical structure of the human skeleton \cite{chen2022hmanet, li2025hyliformer, fish2025geo}. After each looped reconsideration, the model needs to rediscover the spatial geometry tied to the semantics of the sign: where the important cues lie, how small articulations relate to the broader body, and which regions carry meaning in that specific moment. This spatial picture shifts from sign to sign. Sometimes, the message is written in large movements of the arms; sometimes, it exists only in the minute curvature of a finger.

LA-Sign present geometry-aware (GA) alignment as a progressive contrastive objective, regularising the latent space to reflect the intrinsic geometric structure of skeletal motion. As the looped transformer repeatedly revisits its internal representation, this alignment guides refinement by encouraging semantically related patterns to remain close while preserving separation among distinct motions, ensuring that delicate articulatory details are not overshadowed but elevated. Together, looping and GA alignment form a coherent process of refinement, in which each iteration sharpens interpretation and alignment steadily brings meaningful structure into focus.

Our contributions are threefold: 1) We propose a \textbf{looping mechanism for recurrent latent refinement}, which enables the model to repeatedly process skeletal sequences within a compact latent space. This design supports deeper reasoning over temporal dynamics without increasing model parameters. 2) We extend static hyperbolic contrastive regularisation as a recurrent training signal applied across all looping iterations. Unlike static contrastive, which applies hyperbolic alignment as a static regulariser on a single-pass model, we apply the alignment loss progressively at each recurrent step, providing explicit geometric guidance to intermediate latent representations throughout the refinement process. 3) We conduct a comprehensive comparative study of \textbf{multiple looping variants and geometric manifolds}, evaluating their impact on cross-modal representation learning and providing insights into effective architectural designs for skeleton-based ISLR.

% We apply a \textbf{geometry-aware alignment} that leverages hyperbolic space to better preserve the multi-scale relationships within skeletal data. This facilitates more accurate modeling of fine-grained cues, such as finger articulations and facial articulations, alongside the broader body dynamics essential to ISLR.

%Our contribution is summarised as follows: 1) We propose a \textbf{looping mechanism for latent thought}, which enables the model to iteratively refine internal representations within a compact latent space for ISLR. This design supports introspective computation and enhances the ability of the model to "think" over temporally evolving sign sequences without increasing model size. 2) We present a \textbf{geometry-aware alignment} that leverages hyperbolic geometry to better preserve multi-scale relationships within skeletal data. This facilitates more accurate modelling of fine-grained cues, including finger articulations and facial articulations, alongside broader body dynamics essential to ISLR. 3) We conduct a comparative study of \textbf{multiple looping variants and manifold configurations} to evaluate the impact of cross-modal recurrence on representation learning for ISLR.

\section{Related work} 
\subsection{Sign language recognition}
Sign language recognition is a core task in sign language understanding, aiming to recognise individual sign units from video sequences. Early approaches relied on hand-crafted spatio-temporal features to capture signing motion \cite{farhadi2007transfer, fillbrandt2003extraction, ong2005automatic, rastgoo2021sign, starner1995visual}. With the advent of deep learning, RGB-based models became dominant, with convolutional architectures such as TSM \cite{mohaghegh2020advflow}, R3D \cite{hara2017learning}, and I3D \cite{carreira2017quo} widely used to extract visual representations from sign videos \cite{hu2021hand, hu2021global, joze2018ms, koller2018deep, li2020word, li2020transferring, li2022transcribing}. StepNet \cite{shen2024stepnet}, for instance, adopts TSM \cite{lin2019tsm} to model part-level spatio-temporal relationships and achieves strong benchmark performance. Despite their success, RGB-based methods often suffer from static and irrelevant visual cues, such as background clutter, clothing variations, and lighting conditions, which obscure subtle articulatory differences critical to recognition \cite{zuo2023natural}. This has motivated increasing interest in skeleton-based approaches that provide compact, motion-focused representations. NLA-SLR \cite{zuo2023natural} introduces a skeleton-based backbone tailored for sign motion and demonstrates the robustness of skeletal cues, while SAM-v2 \cite{jiang2021sign} further exploits skeleton-derived information to enhance articulation modelling. Most closely related to our geometric alignment component, Geo-Sign \cite{fish2025geo} proposes hyperbolic contrastive regularisation for sign language processing, projecting ST-GCN \cite{yan2018spatial} skeletal features into the Poincaré ball \cite{nickel2017poincare, ganea2018hyperbolic} and aligning them with text embeddings via a geodesic contrastive loss applied as a static objective on a single-pass model. However, most existing skeleton-based methods learn spatio-temporal patterns in a single forward pass and, potentially overlooking fine-grained distinctions that emerge through iterative refinement. In this work, we focus on skeleton-based ISLR and propose LA-Sign, a loop-based transformer that iteratively refines skeletal representations. While Geo-Sign is constrained to the Poincaré ball, we study a variety of manifolds to identify optimal embedding spaces for sign representation. By extending hyperbolic alignment to a recurrent setting and applying it progressively across all looping iterations, LA-Sign captures multi-scale articulatory structure and addresses the fine distinctions fundamental to accurate ISLR.

%However, most existing skeleton-based methods learn spatio-temporal patterns in a single forward pass, potentially overlooking fine-grained distinctions that emerge through iterative refinement. In this work, we focus on skeleton-based ISLR and propose LA-Sign, a loop-based transformer that iteratively refines skeletal representations. By combining iterative depth with GA alignment, LA-Sign captures multi-scale articulatory structure and addresses the fine distinctions fundamental to accurate ISLR.

\subsection{Recurrent looping in latent space}
Early attempts have shown that learning-based models can emulate Turing-complete computation to unfold iteratively for complex problems \cite{dehghani2018universal, perez2019turing, perez2021attention}. The extent to which recurrence is a foundational concept of deep learning is hard to overstate \cite{hopfield1982neural, sutskever2008recurrent}. This underlies recurrent architectures, where fixed-depth blocks are reused with state feedback to refine representations without expanding model size \cite{dehghani2018universal, hutchins2022block, shen2022sliced}. Such recurrent looping enhances features representations \cite{DMDP2019Gelada, wang2020implicit}, which is a valuable property in SLs given its inherent ambiguity and weak linguistic grounding. This mechanism has also proven effective in tasks requiring structured processing, such as code as well as logic, and offers an implicit, efficient alternative to chain-of-thought prompting \cite{saunshi2025reasoning, Geiping2025test}. Though not originally designed for vision-language tasks, this paradigm can transfer to ISLR, where iterative refinement helps adapt attention, capture temporal dynamics, and improve outputs. Together, looped frameworks present a practical and resource-efficient approach to deepening model capacity in SLs.

\subsection{Hyperbolic projection on Riemannian manifolds}
Hyperbolic space, a Riemannian manifold with constant negative curvature, offers an elegant form of embedding hierarchical structures \cite{becigneul2018riemannian}. A key geometric property of hyperbolic space is that its volume increases exponentially with distance from the origin. This property enables tree-like or hierarchical relationships to be captured with lower distortion than in Euclidean space \cite{ganea2018hyperbolic}. This property has motivated work in natural language processing to embed discrete elements and preserve lexical hierarchies \cite{nickel2017poincare}. More recently, hyperbolic embeddings have been applied to continuous domains to model structural patterns beyond Euclidean constraints \cite{khrulkov2020hyperbolic, shimizu2020hyperbolic}. However, such approaches are not designed to handle the dynamic, context-dependent nature of spatial structures in SLs, where the importance of skeletal features shifts over time and across signs. To address this, we project multi-scale skeletal features into hyperbolic space adaptively, allowing the model to capture evolving structural cues and produce more meaningful representations of sign sequences.
 
\section{Methodology}
\begin{figure*}[t]
  \centering
  \includegraphics[width=0.9\linewidth]{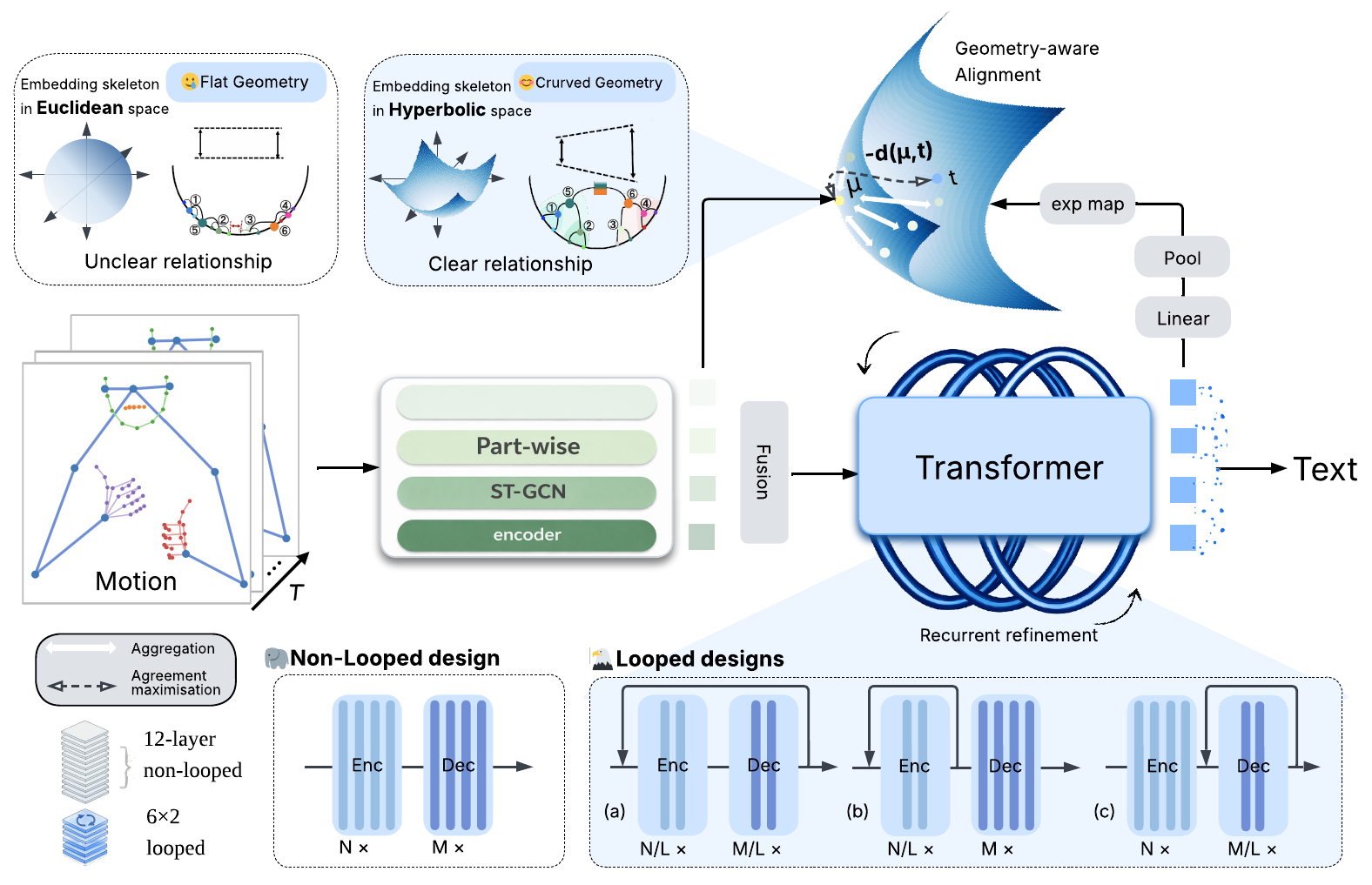}
  \caption{Overview of LA-Sign. Motion sequences are first processed by a part-wise ST-GCN encoder to extract sign features, which are then fed into a looped transformer for recurrent refinement. We study three looping variants: (a) encoder–decoder, (b) encoder-focused, and (c) decoder-focused, to assess how modality interaction patterns affect refinement. To regularise the latent space, we present a GA objective that maps sign and text features onto non-Euclidean manifolds. While traditional geometric approaches apply this alignment as a static, final objective, we integrate it as a recurrent signal computed at each looping iteration. This allows for the continuous minimisation of geodesic distances between paired embeddings throughout the refinement process (see Section \ref{sec:GA}). The refined representation is finally projected to text tokens for recognition.}
  \label{fig:pipeline}
  % https://lucid.app/documents#/documents?folder_id=home
\end{figure*}

LA-Sign realises depth through recurrence rather than deeper stacks (as shown in Figure \ref{fig:pipeline}). A shared looped module revisits the latent states, with each pass sharpening salient motion cues, without increasing the parameter count. In parallel, geometry-aware (GA) alignment shapes the latent space through a contrastive objective, pulling related motion patterns closer while separating distinct articulations. Applied across recurrence, alignment guides refinement toward visually coherent structures that mirror the organisation of signing motion. Together, looping and GA alignment form a unified process in which interpretation gradually comes into focus.

%Sign features are aggregated via the Fréchet mean, while text features are pooled and projected to the corresponding manifold space.

\subsection{Preliminaries}
We train LA-Sign using paired sign sequences and textual labels. The sign inputs consist of skeletal sequences extracted via RTM-Pose \cite{jiang2023rtmpose} from MMPose \cite{sengupta2020mm}, which are partitioned into four anatomical parts: upper body ($b$), face ($f$), left hand ($l$), and right hand ($r$). For each part $p \in \{b,f,l,r\}$, the raw coordinates $S^{p} \in \mathbb{R}^{L \times N^p \times K}$ are independently processed by a part-wise ST-GCN encoder \cite{yan2018spatial} to capture motion dynamics.

These part-wise features $Z^p \in \mathbb{R}^{F \times d_{\mathrm{gcn}}}$ are then concatenated and linearly projected into a shared latent space by the fusion module to form dynamic Euclidean pose embeddings $S \in \mathbb{R}^{F \times d_{\mathrm{mT5}}}$. These embeddings serve as input to the transformer, which is fully fine-tuned from the mT5 base model \cite{xue2020mt5, li2025uni}. 

Here, $F$ denotes the sequence length, $N^p$ is the number of keypoints of part $p$, $K=3$ is the keypoint 2D coordinates and confidence scores, and $d_{\mathrm{gcn}}$ and $d_{\mathrm{mT5}}$ are the hidden dimensions of the ST-GCN and transformer, respectively.

\subsection{Recurrent looping for iterative refinement}
The capability to think before acting often resolves ambiguity and fosters coherent understanding. Recurrence serves as a fundamental learning mechanism, echoing the human tendency to consolidate understanding through experience. This principle finds a parallel in machine learning, where iterative updates via gradient descent progressively refine the parameters of a model. Drawing further inspiration from recurrent activation patterns observed in the human brain while thinking, we introduce a latent recurrent looping mechanism that incrementally refines multimodal representations for ISLR.

\subsection{Design variants of recurrent looping}
While recurrence provides a general mechanism for representation refinement, its impact depends critically on where looping is introduced. In ISLR, visual motion understanding and linguistic decoding play complementary yet distinct roles. Looping can therefore be applied at different stages of the model, leading to qualitatively different refinement behaviours.

\begin{figure}[htbp]
  \centering
  \includegraphics[width=0.6\linewidth]{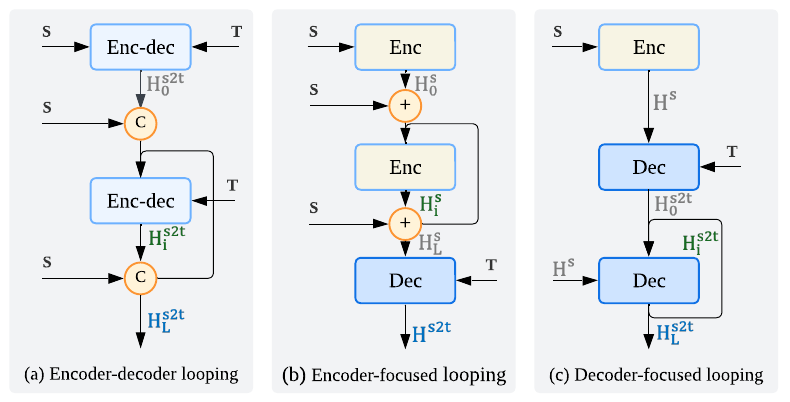} 
  \caption{Architectural details of the three recurrent looping variants. \textbf{(a) Encoder-decoder looping:} the initial sign representation $S$ is concatenated with the previous cross-modal state $H^{s2t}_{i-1}$ before passing through the shared encoder-decoder block. \textbf{(b) Encoder-focused looping:} the visual representation $H^{s}_i$ is iteratively refined via residual updates with $S$. \textbf{(c) Decoder-focused looping:} the encoder processes $S$ into a fixed visual representation $H^s$, which the looped decoder repeatedly references to refine the cross-modal interpretation $H^{s2t}_i$. In all variants, $T$ provides the linguistic context or decoder prefix. Features highlighted in green and blue denote intermediate and final representations, respectively; both are utilised in our GA alignment.}
  \label{fig:loop_variants}
  % https://lucid.app/lucidchart/766e9211-d268-4baf-9758-17df3e625179/edit?beaconFlowId=B56518EE8E3C7D41&invitationId=inv_2f3093af-1bcb-4c87-845f-c8db6cacb66b&page=0_0#
\end{figure}

To investigate these effects, we consider three looped designs (as illustrated in Figure \ref{fig:loop_variants}): (a) encoder– decoder looping, (b) encoder-focused looping, and (c) decoder-focused looping. These variants differ in how the sign representation $S$, the visual sign representation $H^{s}$ by the encoder, and the cross-modal sign-to-text representation $H^{s2t}$ by the decoder are reused across iterations. By comparing these configurations, we aim to disentangle whether iterative refinement is effective when applied jointly to perception and generation, or when concentrated on visual encoding or linguistic decoding alone.

To ensure fair comparison across variants, we balance the number of recurrent steps and the depth of each looped block such that the overall computational depth remains approximately constant. Specifically, when applying $L$ recurrent iterations, the encoder and decoder depths are scaled to $M/L$ and $N/L$, respectively. 

\subsubsection{Encoder-decoder looping}
Encoder-decoder looping applies recurrence to both the encoder and decoder. At each iteration $i$, the previous latent cross-modal representation $H^{s2t}_{i-1}$ is concatenated with the initial sign representation $S$ and re-encoded. During training, the decoder receives linguistic supervision via teacher forcing, where $T$ denotes the embedded text representation provided to the decoder. During inference, $T$ instead corresponds to the autoregressively generated prefix embedding. In both cases, the refined visual features guide the decoder to produce the next output tokens. As formalised in Equation \ref{eq:enc_dec}, this iterative refinement progressively sharpens the model’s interpretation of signing motion through structured cross-modal interaction.

\begin{equation} \label{eq:enc_dec}
    H^{s2t}_i = D\big(E(S \oplus H^{s2t}_{i-1}),\, T\big).
\end{equation} 

\subsubsection{Encoder-focused looping}
Encoder-focused looping applies recurrence exclusively to the encoder, while the decoder remains non-looped. The latent visual representation $H^{s}$ is iteratively refined through repeated encoding, using residual updates with the initial sign representation $S$. The decoder is applied only once after visual refinement is complete, as formalised in Equation \ref{eq:enc}. This design prioritises stabilising skeletal motion representations before alignment with language, allowing us to examine whether improved visual understanding alone can drive recognition.

\begin{equation} \label{eq:enc}
H^{s}_i = E(H^{s}_{i-1}+S), \quad H^{s2t} = D(H^{s}_L, T).
\end{equation} 

\subsubsection{Decoder-focused looping}
Decoder-focused looping employs a standard encoder followed by a recurrently looped decoder. As formalised in Equation \ref{eq:dec}, the visual representation $H^s$ is computed once by the encoder, while the decoder repeatedly reprocesses cross-modal interactions between skeletal features and linguistic context. This configuration evaluates whether iterative refinement at the generation stage can compensate for a single-pass visual encoding by progressively sharpening semantic interpretation.

\begin{equation} \label{eq:dec}
H^s = E(S), H^{s2t}_0 = D(H^s, T\big), \quad H^{s2t}_i = D\big(H^s, H^{s2t}_{i-1}\big).
\end{equation} 

Across recurrent iterations, LA-Sign repeatedly revisits and refines its latent representation. Alignment is applied between $S$ and $H^{s2t}_{i}$ for $i \in \{1, \dots, L\}$ to serve as a complementary training signal. We intentionally omit the initial state ($i=0$) from this alignment, as it precedes the recurrent refinement process; regularising it would prematurely constrain the raw visual encoding before any structural "thought" has occurred. In the specific case of encoder-focused looping, where the decoder is not inherently part of the recurrent loop, a shared decoder is temporarily applied to intermediate states $H^{s}_{i}$ to obtain the cross-modal features $H^{s2t}_{i}$. To stabilise training for the encoder-focused looping, gradients from these intermediate decoding passes are detached; only the final decoder pass at iteration $L$ contributes to the gradient update of the decoder parameters. Applied across all iterations, this alignment encourages semantically related motion patterns to move closer while preserving separation among distinct articulations, providing explicit structural guidance to each refinement cycle.

\subsection{Geometry-aware alignment} \label{sec:GA}
While Euclidean space is the default choice for most neural embeddings, its flat geometry limits its ability to model the multi-scale structure that characterises SL motion. Subtle finger articulations, coordinated hand shapes, facial cues, and broader arm movements coexist across different spatial extents and interact in a non-uniform manner. To move beyond the constraints of flat geometry, LA-Sign incorporates GA alignment, a hyperbolic contrastive objective that regularises the latent space according to the multi-scale organisation of skeletal motion and text semantics. We extend traditional hyperbolic contrastive regularisation from a static, single-pass setting to a recurrent one, applying alignment progressively across all looping iterations to provide explicit geometric guidance to intermediate latent representations.

Hyperbolic geometry \cite{li2025enhancing, liu2025hyperbolic, desai2023hyperbolic, fish2025geo} provides an expressive alternative through its constant negative curvature and exponential volume growth, which allow distances to expand non-linearly as representations move away from the origin. This property supports the coexistence of coarse and fine-grained structures within a single embedding space. Within this geometric family, two models are commonly used: the Poincaré ball, which offers a bounded and conformal view of hyperbolic space, and the Lorentz model, which embeds hyperbolic space as a hyperboloid in Minkowski geometry. While both share equivalent geodesic structure, the Poincaré model provides a more interpretable and intuitive representation, making it suitable for modelling skeletal motion. In LA-Sign, curvature is made adaptive through a learnable scaling factor, enabling the geometric capacity of the space to adjust to the varying structural complexity of different signing sequences.

For completeness, we defer the formal definitions of Euclidean, Poincaré, and Lorentz geometries, along with their distance formulations, to the appendix.

\subsubsection{Adaptive Poincaré ball modelling}
As the recurrent module iteratively refines skeletal motion representations, the GA alignment loss shapes the latent space to ensure that representations reflecting similar motion patterns across scales remain topologically close, while structurally distinct motions are well-separated. This process fosters consistent manifold organisation without introducing extraneous features or disrupting the underlying recurrent computation. Thus, the recurrent loop refines the representations, while GA alignment provides the structural constraints.

To balance expressive power with flexibility, we adopt an adaptive Poincaré ball $\mathbb{B}_{\hat{c}}^d$ with an effective curvature $\hat{c}$. The curvature is defined dynamically as:

\begin{equation}
    \hat{c} = \sigma \cdot c,
\end{equation}

where $c > 0$ represents a fixed base curvature and $\sigma > 0$ is a learnable scaling factor. This formulation allows the geometry to expand or contract dynamically, allocating higher representational capacity where fine-grained distinctions are required and compressing the space when a coarser structure suffices.

The geodesic distance between two points $\mathbf{u}, \mathbf{v} \in \mathbb{B}^d_{\hat{c}}$ is defined as:

\begin{equation}
    d_{\mathbb{B}_{\hat{c}}}(\mathbf{u}, \mathbf{v}) = \frac{2}{\sqrt{\hat{c}}} \tanh^{-1}\!\left( \sqrt{\hat{c}} \left\| (-\mathbf{u}) \oplus_{\hat{c}} \mathbf{v} \right\|_2 \right),
\end{equation}

where $\oplus_{\hat{c}}$ denotes Möbius addition \cite{ganea2018hyperbolic} associated with curvature $\hat{c}$. Euclidean vectors $\mathbf{v}$ in the tangent space $\mathcal{T}_0 \mathbb{B}^d_{\hat{c}}$ are mapped to the hyperbolic manifold via the exponential map at the origin:

\begin{equation}
    \exp^{\hat{c}}_0(\mathbf{v}) = \tanh\!\left( \frac{\sqrt{\hat{c}}\|\mathbf{v}\|_{2}}{2} \right) \frac{\mathbf{v}}{\sqrt{\hat{c}}\|\mathbf{v}\|_{2}}.
\end{equation}

\textbf{Projection into the Adaptive Poincaré Ball}: For each modality $x \in \{s, t\}$, corresponding to the sign and text modalities respectively, the Euclidean features $X$ of the input data are projected into the adaptive Poincaré ball as follows:
\begin{equation}
   \mathcal{H}^x = \exp^{\hat{c}}_0\!\left(\mathbf{W} X \right),
\end{equation}
where $\mathbf{W}$ is a learnable linear projection matrix. Text features are summarised via Euclidean mean pooling prior to projection, whereas sign features are aggregated using the weighted Fréchet mean after projection. This distinction ensures that multi-scale motion patterns are coherently reflected in hyperbolic space, effectively aligning skeletal dynamics with their corresponding linguistic descriptions.

\textbf{Weighted Fréchet Mean Aggregation}: Sign features are aggregated using a weighted Fréchet mean \cite{frechet1948elements, fish2025geo} on the hyperbolic manifold. We optimise the mean via iterative Riemannian updates. More details are provided in the appendix.

\textbf{Riemannian Optimisation}: Hyperbolic embeddings are updated using Riemannian gradients \cite{becigneulriemannian}, while curvature, scaling factors, and temperature settings are optimised in Euclidean space.

%The model employs a hybrid optimisation strategy. Hyperbolic embeddings are updated using Riemannian gradients to preserve the manifold geometry, while curvature parameters, scaling factors, and temperature settings are updated using standard Euclidean optimisers. This scheme ensures stable convergence and allows the geometry to adapt smoothly to the complex, multi-scale structure of sign language motion.

\subsubsection{Contrastive Alignment in Hyperbolic Space}
The recurrent GA alignment is enforced using a hyperbolic contrastive loss at each recurrent iteration $i \in \{1, \dots, L\}$, and aggregated across all iterations as:

\begin{multline}
\mathcal{L}_{GA} = \underbrace{-\log \frac{ \exp\left( -\frac{d_{\mathbb{B},\hat{c}}(\mu_L, \mathcal{H}_L^{t})}{\tau} \right) }{ \sum_{k} \exp\left( -\frac{ d_{\mathbb{B},\hat{c}}(\mu_L, \mathcal{H}_k^{t})}{\tau} + m \cdot \mathbb{I}[k \neq L] \right) }}_{\mathcal{L}_{GA,L} \ (\text{primary})} \\
+ w_{\text{aux}} \sum_{i=1}^{L-1} \underbrace{\left( -\log \frac{ \exp\left( -\frac{d_{\mathbb{B},\hat{c}}(\mu_i, \mathcal{H}_i^{t})}{\tau} \right) }{ \sum_{k} \exp\left( -\frac{ d_{\mathbb{B},\hat{c}}(\mu_i, \mathcal{H}_k^{t})}{\tau} + m \cdot \mathbb{I}[k \neq i] \right) } \right)}_{\mathcal{L}_{GA,i} \ (\text{auxiliary})}
\end{multline}

where $\tau > 0$ is a learnable temperature parameter, $m \ge 0$ is a margin that encourages stricter separation between mismatched pairs, $\mathcal{L}_{GA,L}$ denotes the primary alignment loss at the final iteration $L$, and $w_{\text{aux}}$ controls the contribution of the intermediate auxiliary losses from preceding iterations.

\subsubsection{Language modelling} \label{seq:loop_train_obj}
We apply a \textit{LM} loss at the end of the loop as below:

\begin{equation}
\mathcal{L}_{LM} = - \sum_{n=1}^{N^w} \log p(s_n \mid s_{<n}, S)
\end{equation}

where $s_n$ represents the n-th token, $s_{<n}$ denotes all preceding tokens in the sentence $S$, and $L_{LM}$ is optimised by AdamW gradient descent.

%Together, looping and GA alignment form a unified process of refinement and organisation. Training proceeds by jointly optimising a language modelling (\textbf{LM}) loss and a \textbf{GA} alignment loss that encourages structural and semantic coherence across iterations.

\subsection{Training objective} 
The joint training objective combines the \textit{LM} loss and the \textit{GA} loss using a learnable scalar $\alpha$:

\begin{equation}
    \mathcal{L}_{joint} = \alpha \mathcal{L}_{LM} + (1 - \alpha)\mathcal{L}_{GA} 
\end{equation}

where $w_{\text{aux}}$ is a hyperparameter that controls the contribution of the intermediate auxiliary losses, which is set to 0.1 to ensure that intermediate alignments guide the representation without overshadowing the primary alignment at the final iteration. Specifically, $\mathcal{L}_{GA_L}$ denotes the primary GA alignment loss computed at the final recurrent iteration $L$, while the summation term accumulates the auxiliary alignment losses from all preceding iterations (from $i=1$ to $L-1$). By enforcing geometric alignment at every step of the loop, these auxiliary losses provide explicit structural guidance to the intermediate latent thoughts, preventing the representations from degrading during the recurrent process. The scalar $\alpha$ is adjusted based on agreement feedback to balance representative and generative objectives during training through the GA loss and the LM loss. This dynamic scheduling allows the model to benefit from different loss emphases at different training stages. This module encourages semantically structured alignment between modalities and grounds the reflection processing of the model within a hyperbolically aware space.

\section{Experiments}
We evaluate our method on WLASL and MSASL benchmarks, two widely adopted and representative datasets of over 40000 videos. Model performance is assessed using per-class (P-C) and per-instance (P-I) Top-1 accuracy. All models strictly follow the architectural designs described in this paper and adopt the same training stages and dataset configuration as the pose-only setting of \cite{li2025uni}. 

\subsection{Comparison with state-of-the-art methods}
We evaluate LA-Sign against state-of-the-art (SOTA) methods, as summarised in Table \ref{tab:exp} (d).

WLASL. LA-Sign achieves new state-of-the-art performance on WLASL. On WLASL2000, it reaches 64.73\% P-I and 62.41\% P-C. It outperforms Geo-Sign \cite{fish2025geo}, the method upon which our geometric alignment is extended, by +1.09\% P-I and +0.52\% P-C, despite operating with a 6-layer shared encoder-decoder against Geo-Sign's full 12-layer mT5 model, validating the benefit of applying hyperbolic alignment recurrently rather than as a static objective. On WLASL300, LA-Sign improves over NLA-SLR by +1.68\% P-I and +2.94\% P-C, demonstrating strong generalisation.

MSASL. LA-Sign also outperforms existing approaches on MSASL. On MSASL1000, it achieves 79.68\% P-I, improving over Uni-Sign by +1.52\%. On MSASL200, it obtains 91.40\% P-I and 94.10\% P-C, exceeding CCL-SLR by +1.19\% and +3.24\%, respectively. These consistent gains confirm the effectiveness of the proposed framework.

\begin{table*}[ht]
    \centering
    \caption{Performance analysis and benchmark comparison. (a) Base vs. Loop architectures. (b) Looping designs. (c) Geometric manifolds. (d) Comparison with SOTA methods on WLASL and MSASL. Configurations in (a) are denoted as $(U \times L)$ ($U$ unique layers, $L$ recurrent depthes) keeping effective computational depth constant. \colorbox{darkb}{\textbf{Best}} and \colorbox{lightb}{second-best} results are highlighted. $*$ indicates additional modalities. $\dag$ denotes multi-crop inference. $\ddag$ denotes optical flow fusion.}
    \label{tab:exp}
    \renewcommand{\arraystretch}{1.1}
    % --- Abalation ---
    \begin{minipage}[t]{0.29\textwidth}
        \vspace{0pt}
        % (a) Base and Loop
        \centering
        {\small (a) Base and Loop.} \\
        \resizebox{\linewidth}{!}{ 
            \begin{tabular}{lll}
            \toprule
            Method & P-I & P-C \\ 
            \midrule
            Base ($12 \times 1$) & \cellcolor{lightb}63.55 & \cellcolor{lightb}61.35 \\ 
            Base ($2 \times 1$) & 0.13 & 0.13 \\ 
            Loop ($2 \times 6$) & 33.81 & 33.70 \\ 
            Base ($4 \times 1$) & 7.91 & 7.57 \\ 
            Loop ($4 \times 3$) & 35.44 & 36.87 \\ 
            Base ($6 \times 1$) & 23.17 & 20.12 \\ 
            Loop ($6 \times 2$) & \cellcolor{darkb}\textbf{64.73} & \cellcolor{darkb}\textbf{62.41} \\ 
            \bottomrule
            \end{tabular}
        }
        
        \vspace{0.5em}
        
        % (b) Looped designs
        \centering
        {\small (b) Looped designs.} \\
        \resizebox{\linewidth}{!}{
            \begin{tabular}{lll}
            \toprule
            Design & P-I & P-C \\ 
            \midrule
            encoder-decoder & \cellcolor{darkb}\textbf{64.73} & \cellcolor{darkb}\textbf{62.41} \\ 
            decoder & 63.54 & \cellcolor{lightb}61.40 \\ 
            encoder & \cellcolor{lightb}63.73 & 61.31 \\ 
            \bottomrule
            \end{tabular}
        }
        
        \vspace{0.5em}
        
        % (c) Manifold
        \centering
        {\small (c) Manifold..} \\
        \resizebox{\linewidth}{!}{
            \begin{tabular}{lll}
            \toprule
            Manifold & P-I & P-C \\ 
            \midrule
            Euclidean (Baseline) & 63.93 & 61.71 \\ 
            Poincaré & 64.62 & 61.76 \\ 
            Lorentz & \cellcolor{lightb}64.66 & \cellcolor{lightb}62.31 \\ 
            Adaptive Poincaré & \cellcolor{darkb}\textbf{64.73} & \cellcolor{darkb}\textbf{62.41} \\ 
            Adaptive Lorentz & 64.43 & 61.53 \\ 
            \bottomrule
            \end{tabular}
        }
    \end{minipage}
    \hfill
    % --- SOTA ---
    \begin{minipage}[t]{0.7\textwidth}
        \vspace{0pt}
        \centering
        {\small (d) Performance comparison on various benchmarks.} \\
        \renewcommand{\arraystretch}{1.24}
        \resizebox{\linewidth}{!}{
            \begin{tabular}{lcccccccc}
            \toprule
            \multirow{2}{*}{Method} & \multicolumn{2}{c}{WLASL2000} & \multicolumn{2}{c}{WLASL300} & \multicolumn{2}{c}{MSASL1000} & \multicolumn{2}{c}{MSASL200} \\ 
            \cmidrule(lr){2-3} \cmidrule(lr){4-5} \cmidrule(lr){6-7}
            \cmidrule(lr){8-9}
             & P-I & P-C & P-I & P-C & P-I & P-C & P-I & P-C \\ 
            \midrule
            ST-GCN \cite{yan2018spatial} & 34.40 & 32.53 & 44.46 & 45.29 & 36.03 & 32.32 & 52.91 & 54.20 \\ 
            I3D \cite{carreira2017quo} & 32.48 & - & 56.14 & - & - & 57.69 & - & 81.97 \\ 
            TCK \cite{li2020transferring} & - & - & 68.56 & 68.75 & - & - & 80.31 & 81.14 \\ 
            StepNet \cite{shen2024stepnet} & 56.89 & 54.54 & 74.70 & 75.32 & - & - & - & - \\ 
            StepNet\ddag \cite{shen2024stepnet} & 61.17 & 58.43 & - & - & - & - & - & - \\ 
            HMA \cite{hu2021hand} & 51.39 & 48.75 & - & - & 69.39 & 66.54 & 85.21 & 86.09 \\ 
            BEST \cite{zhao2023best} & 54.59 & 52.12 & 75.60 & 76.12 & 71.21 & 68.24 & 86.83 & 87.45 \\ 
            SignBERT \cite{hu2021signbert} & 54.69 & 52.08 & 74.40 & 75.27 & 71.24 & 67.96 & 86.98 & 87.62 \\ 
            SignBERT+ \cite{hu2023signbert+} & 55.59 & 53.33 & 78.44 & 79.12 & 73.71 & 70.77 & 88.08 & 88.62 \\ 
            SAM*\dag \cite{jiang2021skeleton}  & 58.73 & 55.93 & - & - & - & - & - & - \\ 
            SAM-v2*\dag \cite{jiang2021sign} & 59.39 & 56.63 & - & - & - & - & - & - \\ 
            NLA-SLR \cite{zuo2023natural} & 61.05 & 58.05 & 86.23 & 86.67 & 72.56 & 69.86 & 88.74 & 89.23 \\ 
            NLA-SLR\dag \cite{zuo2023natural} & 61.26 & 58.31 & \cellcolor{lightb}86.98 & \cellcolor{lightb}87.33 & 73.80 & 70.95 & 89.48 & 89.86 \\ 
            CCL-SLR \cite{wu2025cross} & 62.20 & 58.80 & 86.28 & 86.81 & 77.71 & 76.27 & \cellcolor{lightb}90.21 & \cellcolor{lightb}90.86 \\ 
            Uni-Sign \cite{liuni} & 63.52 & 61.32 & - & - & \cellcolor{lightb}78.16 & \cellcolor{lightb}76.97 & - & - \\ 
            Geo-Sign \cite{fish2025geo}  & \cellcolor{lightb}63.64 & \cellcolor{lightb}61.89 & - & - & - & - & - & - \\ 
            LA-Sign (ours) & \cellcolor{darkb}\textbf{64.73} & \cellcolor{darkb}\textbf{62.41} & \cellcolor{darkb}\textbf{88.66} & \cellcolor{darkb}\textbf{90.27} & \cellcolor{darkb}\textbf{79.68} & \cellcolor{darkb}\textbf{79.64} & \cellcolor{darkb}\textbf{91.40} & \cellcolor{darkb}\textbf{94.10} \\ 
            \bottomrule
            \end{tabular}
        }
    \end{minipage}
\end{table*}

\subsection{On the power of the looped Transformer} \label{exp:base_vs_loop}
To evaluate recurrent looping, we compare standard layer-stacking ("Base") against our encoder-decoder "Loop" design, denoted as ($U \times L$) for $U$ unique layers and $L$ recurrent passes. Keeping effective computational depth constant, Table \ref{tab:exp} (a) shows that reducing unique layers in the baseline severely degrades performance: halving the 12-layer Base ($12 \times 1$) drops P-I from 63.55 to 23.17, while Base ($2 \times 1$) collapses entirely (0.13 P-I). Introducing the loop mechanism substantially recovers this capacity. For instance, Loop ($4 \times 3$) and ($2 \times 6$) boost P-I to 35.44 and 33.81, respectively, rescuing performance from their near-zero non-looped counterparts. Most notably, Loop ($6 \times 2$) achieves the highest overall scores (64.73 P-I, 62.41 P-C), surpassing the full 12-layer Base model. This demonstrates that a 6-layer architecture with a single recurrent iteration can exceed the representational power of a deeper network. We attribute this to the regularising effect of weight sharing and iterative feature alignment, which refines representations more efficiently than distinct, uncoupled layers. A qualitative visualisation of this evolution is provided in the Appendix.

\subsection{Impact of the design variants of recurrent looping} \label{exp:recur_depth}
To determine the optimal placement of the recurrent loop, we evaluated applying it exclusively to the encoder (focusing on visual perception), exclusively to the decoder (focusing on linguistic generation), and jointly across both. As shown in Table \ref{tab:exp} (b), isolating recurrent refinement to a single component yields suboptimal results. Looping only the encoder achieves 63.73 P-I and 61.31 P-C, while looping only the decoder results in 63.54 P-I and 61.40 P-C. However, coupling the loop across the full encoder-decoder pipeline unlocks the model's peak performance, reaching 64.73 P-I and 62.41 P-C. This notable gap underscores the necessity of jointly iterating feature extraction and sequence generation. Synchronizing the recurrent depth allows the model to progressively refine visual source representations in tandem with their cross-modal target alignments, effectively bridging the semantic gap between continuous skeletal motion and discrete text. By contrast, restricting the loop to a single component creates a representational bottleneck: one side of the network engages in deep, iterative reasoning while the other remains a shallow, single-pass operation, leading to a structural mismatch that hinders the overall recognition capacity.

\subsection{Impact of manifold geometry}
To investigate the influence of the underlying geometric space on representation learning, we evaluated the model's performance across different manifolds, including the standard Euclidean space and two variations of hyperbolic space (Poincaré and Lorentz). We also tested adaptive variants where the curvature of the manifold is jointly learned during training. As demonstrated in Table \ref{tab:exp} (c), transitioning from a flat Euclidean space to a hyperbolic geometry yields immediate improvements. The Euclidean baseline achieves 63.93 P-I and 61.71 P-C. By contrast, embedding features in the static Poincaré ball and Lorentz model elevates the P-I to 64.62 and 64.66, respectively. This confirms that hyperbolic spaces are better suited for capturing the complex, potentially hierarchical structures inherent in continuous sign language data. The introduction of adaptive curvature further highlights the importance of geometric flexibility. The Adaptive Poincaré manifold achieves the highest overall performance in the ablation study, reaching 64.73 P-I and 62.41 P-C. This indicates that allowing the network to dynamically optimize the curvature of the Poincaré ball to match the data distribution provides a superior inductive bias. Interestingly, the Adaptive Lorentz variant experiences a slight degradation compared to its static counterpart (dropping to 64.43 P-I). We attribute this to potential optimization instabilities or numerical precision challenges that can arise when dynamically adjusting the curvature parameters within the Lorentz constraint formulation.

\subsection{Geometric structure of learned embedding space}
\begin{figure}[htbp]
  \centering
  \includegraphics[width=.75\linewidth]{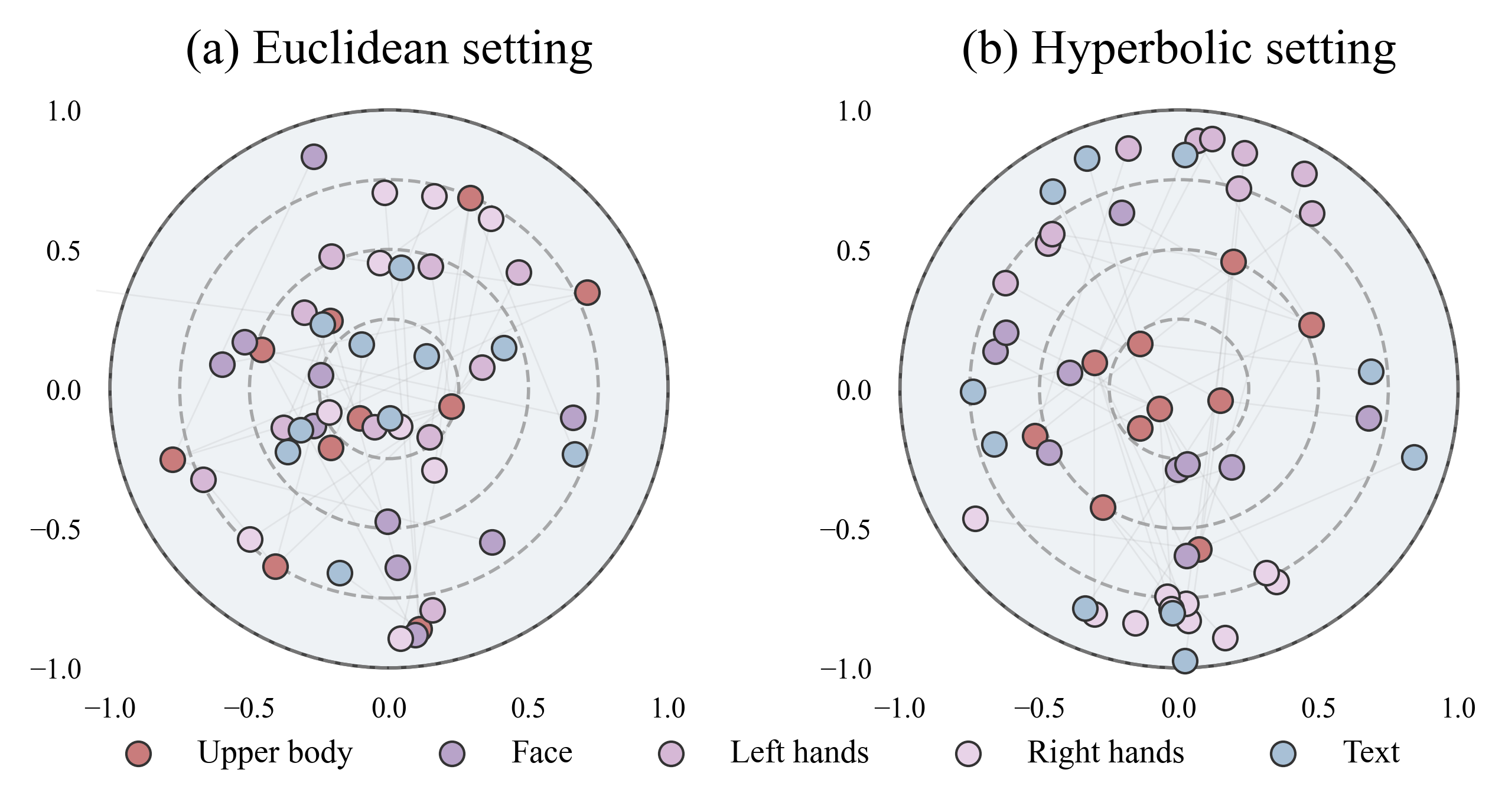}
  \caption{UMAP visualisations of learned embeddings. Compared to the Euclidean baseline (a), the hyperbolic space (b) exhibits clearer radial organisation and improved semantic separation.}
  \label{fig:embed_compare}
\end{figure}

To understand the effect of geometric modelling, we visualise the learned embeddings via UMAP \cite{healy2024uniform}. For the hyperbolic case, embeddings are first log-mapped to the tangent space at the origin. As shown in Figure \ref{fig:embed_compare} (a), Euclidean embeddings lack clear semantic differentiation, which uses the same contrastive alignment objective but without hyperbolic projection, corresponding to curvature c = 0. In contrast, the hyperbolic setting (Figure \ref{fig:embed_compare} (b)) produces a relatively structured radial organisation. Semantically distinct components occupy different radial regions of the Poincaré disk, demonstrating the intrinsic capacity of hyperbolic geometry to encode hierarchical structures and achieve improved representational efficiency.

\section{Conclusion}
We presented LA-Sign, a looped transformer framework with GA alignment for skeleton-based ISLR. By replacing deeper layer stacking with recurrent latent refinement, LA-Sign enables progressive interpretation of signing motion under shared parameters. Building on traditional static hyperbolic contrastive regularisation, we extend GA alignment from a static, single-pass objective to a recurrent training signal applied progressively across all looping iterations, structuring the latent space through adaptive hyperbolic modelling and promoting coherent organisation of multi-scale articulatory cues. Through extensive experiments, we demonstrated that encoder–decoder looping provides the most effective refinement pathway, and that adaptive Poincaré geometry offers superior inductive bias for capturing the structured nature of sign language motion. Empirical results on WLASL and MSASL benchmarks confirm consistent improvements over strong baselines and prior state-of-the-art methods, including Geo-Sign, validating that recurrent geometric alignment yields stronger representations than its static counterpart. Our findings suggest that iterative latent reasoning and geometry-aware representation learning form a powerful combination for sign language understanding. Beyond ISLR, we believe this paradigm offers a promising direction for other vision-language tasks that require structured interpretation under limited supervision.


% \subsubsection{Contrastive alignment in hyperbolic space}
% GA alignment is enforced using a hyperbolic contrastive loss:

% \begin{equation}
%     \mathcal{L}_{GA} = -\log \frac{ \exp\!\left( -\frac{d_{\mathbb{B}_{\hat{c}}}(\mu_i, \mathcal{H}_i^{t})}{\tau} \right) }{ \sum_{k} \exp\!\left( -\frac{ d_{\mathbb{B}_{\hat{c}}}(\mu_i, \mathcal{H}_k^{t})}{\tau} + m \cdot \mathbb{I}[k \neq i]  \right) },
% \end{equation}

% where $\tau > 0$ is a learnable temperature parameter and $m \ge 0$ is a margin that encourages stricter separation between mismatched pairs.
\appendix
\section{Implementation Details}
LA-Sign is implemented in PyTorch and trained using DeepSpeed for distributed optimisation. The training follows a two-stage protocol. The first stage consists of a unified pre-training phase on YouTube-ASL \cite{uthus2023youtube} following the first stage of Uni-Sign \cite{li2025uni}, with the exception that model depths are modified to align with the configurations in Table 1 (a) of the main paper. For downstream tasks or specific benchmark evaluation, models are trained for 25 epochs with a batch size of 8. We utilise the AdamW optimiser with a learning rate of 1.00E-04 and a cosine-annealed scheduler with 5 warmup epochs. To handle the hyperbolic alignment components, manifold parameters are optimised via Riemannian Adam at a learning rate of 1.00E-02, while all other parameters follow the standard Euclidean optimiser. We set the auxiliary alignment weight $w_{\text{aux}} = 0.1$ to provide intermediate geometric supervision without overshadowing the primary signal at the final iteration. The loss blending factor $\alpha$ is initialised at 0.8 and dynamically adjusted throughout training. All experiments are conducted on NVIDIA A100 GPUs.

\section{Skeleton extraction and keypoint configuration}
Skeleton sequences are extracted using RTM-Pose \cite{jiang2023rtmpose} from the MMPose framework \cite{sengupta2020mm}. Each frame contains keypoints from four body parts: left hand, right hand, upper body, and face. Specifically, 21 keypoints are used for each hand, 9 for the body, and 18 for the face, resulting in a total of 69 keypoints per frame. These body parts are processed separately to preserve articulatory structure and reduce interference across regions with different motion characteristics.

\section{Geometric models}
\subsection{Euclidean geometry}
Most neural representations operate in Euclidean space $\mathbb{R}^d$, where distances are measured using the standard $L_2$ norm:
\begin{equation}
d(\mathbf{x}, \mathbf{y}) = ||\mathbf{x} - \mathbf{y}||_2 .
\end{equation}
Here, $\mathbf{x}$, $\mathbf{y}$ $\in \mathbb{R}^d$ denote two feature vectors, and the distance corresponds to the straight-line length between them. This metric assumes a flat geometry in which distances grow linearly with displacement. While Euclidean geometry is simple and widely adopted, its linear growth limits its ability to represent multi-scale relationships, where different patterns may require non-uniform separation to remain distinguishable.

\subsection{Poincaré ball model}
The Poincaré ball $\mathbb{B}^d$ represents hyperbolic space within a bounded unit ball of $d$-dimensional. 
\begin{equation}
\mathbb{B}^d = \{ \mathbf{x} \in \mathbb{R}^d \mid ||\mathbf{x}|| < 1 \}
\end{equation}
where $||\cdot||$ denotes the Euclidean norm. The geodesic distance between two points $\mathbf{x}, \mathbf{y} \in \mathbb{B}^d$ is defined as:
\begin{equation}
d_{\mathbb{B}}(\mathbf{x}, \mathbf{y})
= \mathrm{arcosh} \left(
1 + 2\frac{||\mathbf{x} - \mathbf{y}||^2}
{(1 - ||\mathbf{x}||^2)(1 - ||\mathbf{y}||^2)}
\right).
\end{equation}
This distance measures the length of the shortest geodesic curve connecting $\mathbf{x}$ and $\mathbf{y}$ within the hyperbolic manifold. As points approach the boundary of the ball, distances increase exponentially, even for small Euclidean displacements. This property allows the Poincaré ball to naturally accommodate representations that differ across scales, supporting the coexistence of coarse global patterns and fine-grained local details within a single embedding space.

\subsection{Lorentz model}
The Lorentz model represents hyperbolic space as the upper sheet of a two-sheeted hyperboloid in the Minkowski space $\mathbb{R}^{d+1}$. The manifold is defined as:
\begin{equation}
\mathbb{H}^d = { \mathbf{x} \in \mathbb{R}^{d+1} : \langle \mathbf{x}, \mathbf{x} \rangle_{\mathcal{L}} = -1, , x_0 > 0 },
\end{equation}
where $\langle \cdot, \cdot \rangle_{\mathcal{L}}$ denotes the Lorentzian inner product. For two points $\mathbf{x}, \mathbf{y} \in \mathbb{R}^{d+1}$, this inner product is defined as:
\begin{equation}
\langle \mathbf{x}, \mathbf{y} \rangle_{\mathcal{L}} = -x_0 y_0 + \sum_{i=1}^{d} x_i y_i.
\end{equation}
The geodesic distance between two points $\mathbf{x}, \mathbf{y} \in \mathbb{H}^d$ is given by:
\begin{equation}
d_{\mathcal{L}}(\mathbf{x}, \mathbf{y}) = \operatorname{arccosh}\left( - \langle \mathbf{x}, \mathbf{y} \rangle_{\mathcal{L}} \right).
\end{equation}
Since the points lie on the forward sheet of the hyperboloid, the inner product $\langle \mathbf{x}, \mathbf{y} \rangle_{\mathcal{L}}$ is always less than or equal to $-1$, ensuring the argument of $\operatorname{arccosh}$ is greater than or equal to $1$.

This formulation provides a numerically stable way to compute hyperbolic distances and is well-suited for gradient-based optimisation. While the Lorentz model is mathematically equivalent to the Poincaré ball in terms of geodesic structure, its coordinates are less intuitive to interpret visually, especially for skeletal motion, where bounded and spatially interpretable representations are often preferred.

\section{Weighted Fréchet mean aggregation}
To aggregate the projected sign features into a unified representation, standard linear averaging is insufficient as it disregards the curvature of the hyperbolic manifold. Instead, we employ the weighted Fréchet mean $\boldsymbol{\mu}$, which generalises the centroid to Riemannian manifolds by minimising the weighted sum of squared geodesic distances. Given a sequence of projected features $\mathcal{H}^s = \{ \mathbf{h}_1, \dots, \mathbf{h}_T \} \subset \mathbb{B}_{\hat{c}}^d$ and normalized attention weights $\tilde{w}_t$, the objective is defined as:
\begin{equation}
\boldsymbol{\mu} = \operatorname*{argmin}{\mathbf{z} \in \mathbb{B}{\hat{c}}^d} \sum_{t=1}^{T} \tilde{w}t d^2{\mathbb{B}{\hat{c}}}(\mathbf{z}, \mathbf{h}t).
\end{equation}
Since a closed-form solution is intractable in hyperbolic space, we solve this using an iterative Riemannian gradient descent algorithm. Initializing $\boldsymbol{\mu}^{(0)} = \mathbf{h}_1$, we iteratively map the points to the tangent space of the current estimate, compute the weighted average, and project the result back to the manifold. The update rule at step $k$ is:
\begin{equation}
\boldsymbol{\mu}^{(k+1)} = \exp{\boldsymbol{\mu}^{(k)}}^{\hat{c}} \left( \sum{t=1}^{T} \tilde{w}t \log{\boldsymbol{\mu}^{(k)}}^{\hat{c}}(\mathbf{h}t) \right).
\end{equation}
Here, $\log{\boldsymbol{\mu}}^{\hat{c}}$ maps points from the manifold to the tangent space $\mathcal{T}_{\boldsymbol{\mu}} \mathbb{B}_{\hat{c}}^d$, and $\exp_{\boldsymbol{\mu}}^{\hat{c}}$ maps the updated tangent vector back to the manifold, ensuring the final representation remains faithful to the hyperbolic geometry while accurately summarizing multi-scale motion dynamics.

\section{Additional experiments}
We present additional experiments to support our work in this section.

\subsection{Evolution across loop iterations}
\begin{figure}[htbp]
  \centering
  \includegraphics[width=1\linewidth]{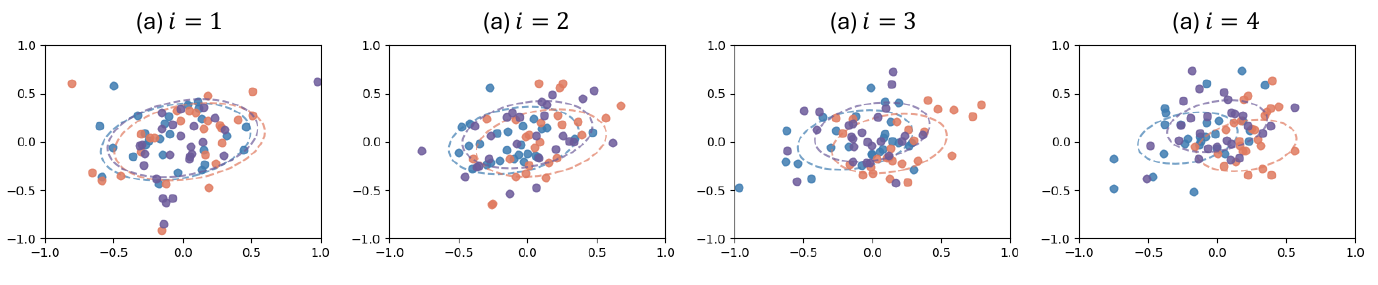} 
  \caption{Representation refinement across loop iterations. We visualise the embedding distributions of three sign classes (“before”, “chair”, and “go”) across successive loop iterations. At early iterations (i = 1), the embeddings exhibit substantial overlap, indicating weak semantic separation. As the number of loops increases, the representations progressively form clearer cluster structures, with reduced inter-class overlap. This behaviour suggests that iterative looping refines the latent representation and enhances semantic organisation in the embedding space.}
  \label{fig:loop_evolution}
\end{figure}

To better understand how the proposed looping mechanism influences representation learning, we visualise the evolution of the embedding space across successive loop iterations. Specifically, we project the learned representations of three sign classes (\textit{before}, \textit{chair}, and \textit{go}) into a two-dimensional space and examine how their distributions change as the number of refinement loops increases. Specifically, the learned sign embeddings are projected into a two-dimensional space using t-SNE for visualisation. Figure \ref{fig:loop_evolution} illustrates the embedding distributions at different loop iterations by using the Loop (4 × 3) model mentioned in the main paper. At the initial stage ($i=1$), the representations are largely mixed, with substantial overlap between classes. This indicates that the early-stage features produced by the model do not yet capture sufficiently discriminative semantic structures. 

As the refinement process proceeds ($i=2$), the clusters begin to exhibit weak structural separation. Although the embeddings still overlap considerably, early signs of class-wise grouping emerge, suggesting that the iterative update starts to organise the latent space. With additional refinement ($i=3$), the cluster structures become more stable and the separation between classes gradually increases. The embeddings of different sign categories begin to occupy distinct regions of the latent space, indicating improved semantic alignment between motion patterns and their corresponding linguistic representations.

Finally, at $i=4$, the clusters become more clearly defined and inter-class overlap is further reduced. This progressive structuring of the embedding space demonstrates that the looping mechanism effectively refines latent representations over iterations. Overall, these observations support the hypothesis that iterative looping acts as a representation refinement process. By repeatedly revisiting the latent features, the model gradually improves both intra-class compactness and inter-class separability, leading to more structured and semantically meaningful embeddings.

\begin{table}[ht]
    \centering
    \definecolor{lightb}{RGB}{225,240,255}
    \definecolor{darkb}{RGB}{190,220,255}
    \renewcommand{\arraystretch}{1.15}
    \caption{Additional ablation studies on geometry-aware alignment and latent refinement mechanisms. 
    (a) Impact of initial curvature in the adaptive Poincaré ball. 
    (b) Impact of initial projection scale. 
    (c) Effect of injecting auxiliary features during looping. 
    (d) Comparison of different input fusion strategies for recurrent refinement. 
    Best and second-best results are highlighted. 
    Results show that adaptive hyperbolic geometry consistently improves recognition, while excessive external injection degrades performance. 
    Concatenation-based fusion and learnable geometric parameters yield the most stable and accurate representations, highlighting the importance of preserving latent continuity during iterative refinement.}
    
    % ---------- Curvature and Scale ----------
    \begin{subtable}[t]{0.48\linewidth}
        \centering
        \caption{Impact of Initial Curvature}
        \begin{tabular}{ll|ll}
        \hline
        Curvature & Learn & P-I & P-C \\
        \hline
        0.01 & No & 63.80 & 61.73 \\
        0.10 & No & 64.06 & 61.82 \\
        1.50 & No & \cellcolor{lightb}64.50 & 62.01 \\
        5.00 & No & 64.22 & \cellcolor{lightb}62.05 \\
        1.50 & Yes & \cellcolor{darkb}\textbf{64.73} & \cellcolor{darkb}\textbf{62.41} \\
        \hline
        \end{tabular}
        \end{subtable}
        \hfill
        \begin{subtable}[t]{0.48\linewidth}
        \centering
        \caption{Impact of Initial Scale}
        \begin{tabular}{ll|ll}
        \hline
        Scale & Learn & P-I & P-C \\
        \hline
        1.00 & No & 63.94 & 61.54 \\
        1.50 & No & 64.29 & 61.86 \\
        2.00 & No & \cellcolor{lightb}64.47 & \cellcolor{lightb}62.05 \\
        2.50 & No & 64.12 & 61.76 \\
        2.00 & Yes & \cellcolor{darkb}\textbf{64.73} & \cellcolor{darkb}\textbf{62.41} \\
        \hline
        \end{tabular}
    \end{subtable}

    \vspace{0.8em}

    % ---------- Extract injection and input injection ----------
    \begin{subtable}[t]{0.48\linewidth}
        \centering
        \caption{Extra Feature Injection}
        \begin{tabular}{l|ll}
        \hline
        Method & P-I & P-C \\
        \hline
        Noise (internal) & 63.35 & 60.94 \\
        Noise (initial) & \cellcolor{lightb}64.21 & \cellcolor{lightb}62.08 \\
        Temporal (initial) & 63.92 & 61.91 \\
        None & \cellcolor{darkb}\textbf{64.73} & \cellcolor{darkb}\textbf{62.41} \\
        \hline
        \end{tabular}
        \end{subtable}
        \hfill
        \begin{subtable}[t]{0.48\linewidth}
        \centering
        \caption{Input injection}
        \begin{tabular}{l|ll}
        \hline
        Strategy & P-I & P-C \\
        \hline
        Addition & \cellcolor{lightb}64.25 & \cellcolor{lightb}62.00 \\
        Softmax fusion & 62.95 & 60.93 \\
        Temporal injection & 63.99 & 61.73 \\
        Concatenation & \cellcolor{darkb}\textbf{64.73} & \cellcolor{darkb}\textbf{62.41} \\
        \hline
        \end{tabular}
    \end{subtable}
    
    \vspace{0.8em}
    
\label{tab:add_exp}
\end{table}

\subsection{Impact of initial curvature}
This study evaluates how the choice of initial hyperbolic curvature affects geometry-aware alignment. Fixed curvature values already outperform Euclidean baselines, with moderate curvature ($c=1.5$) providing the strongest static performance. However, enabling curvature to be learnable produces the best overall result.

This behaviour reflects the heterogeneous nature of sign language motion: different signs exhibit varying degrees of hierarchical structure, ranging from subtle finger articulation to large-scale arm movements. A fixed curvature imposes a uniform geometric bias across all samples, whereas adaptive curvature allows the model to dynamically allocate representational capacity based on motion complexity. This confirms that geometry should not be treated as a static prior but as an optimisable component co-evolving with latent refinement.

\subsection{Impact of initial scale}
We further investigate the influence of projection scale, which controls how Euclidean features are mapped into hyperbolic space. Moderate fixed scaling improves performance, with scale $2.0$ yielding the strongest non-learnable result. However, as in curvature, allowing the scale parameter to be learnable consistently achieves superior accuracy.

This suggests that the effective spread of embeddings within hyperbolic space is task-dependent and evolves during training. Learnable scaling enables the model to balance local discrimination against global organisation, allowing fine-grained articulations and broader motion patterns to coexist without distortion. Together with adaptive curvature, this establishes flexible geometry as a key contributor to robust multi-scale alignment.

\subsection{Extra feature injection}
We examine whether injecting auxiliary signals during recurrent looping improves representation learning. While adding random noise or temporal features yields marginal gains in some cases, the highest performance is achieved when no external features are injected.

This indicates that LA-Sign benefits primarily from internally driven refinement rather than external perturbations. Introducing additional signals disrupts latent continuity across loops, weakening the model’s ability to consolidate representations over iterations. The result supports our design choice: looping should function as a self-consistent reflective process, where refinement emerges from revisiting latent states rather than from artificially introduced variation.

\subsection{Input injection strategy}
Different injection strategies are evaluated for combining initial sign embeddings with looped latent states. Concatenation consistently outperforms additive and attention-based injection.

This outcome highlights the importance of preserving explicit feature channels across iterations. Additive or softmax-weighted injection compresses information prematurely, whereas concatenation maintains separable pathways for original and refined representations. This structural preservation is crucial for iterative refinement, allowing the model to revisit raw motion cues while progressively incorporating higher-level abstractions.

\bibliographystyle{splncs04}
\bibliography{main.bib}
\end{document}